\documentclass[letterpaper]{article} % DO NOT CHANGE THIS
\usepackage{aaai2026}  % DO NOT CHANGE THIS

\usepackage{times}  % DO NOT CHANGE THIS
\usepackage{helvet}  % DO NOT CHANGE THIS
\usepackage{courier}  % DO NOT CHANGE THIS
\usepackage[hyphens]{url}  % DO NOT CHANGE THIS
\usepackage{graphicx} % DO NOT CHANGE THIS
\usepackage{natbib}  % DO NOT CHANGE THIS AND DO NOT ADD ANY OPTIONS TO IT
\usepackage{caption} % DO NOT CHANGE THIS AND DO NOT ADD ANY OPTIONS TO IT
\usepackage{algorithm}
\usepackage{algorithmic}

\usepackage{newfloat}
\usepackage{listings}
\DeclareCaptionStyle{ruled}{labelfont=normalfont,labelsep=colon,strut=off} % DO NOT CHANGE THIS
\floatstyle{ruled}
\newfloat{listing}{tb}{lst}{}
\floatname{listing}{Listing}
\usepackage{adjustbox}
\usepackage{xcolor}
\usepackage{textcomp}

\title{Decoding Emotion in the Deep: A Systematic Study of How LLMs Represent, Retain, and Express Emotion}
\author{
    Jingxiang Zhang\textsuperscript{\rm 1},
    Lujia Zhong\textsuperscript{\rm 1}
}

\affiliations{
    \textsuperscript{\rm 1}University of Southern California\\
    jzhang92@usc.edu, lujiazho@usc.edu
}

\begin{document}

\makeatletter
\def\copyright@on{F}
\makeatother
\maketitle

\begin{abstract}
Large Language Models (LLMs) are increasingly expected to navigate the nuances of human emotion. While research confirms that LLMs can simulate emotional intelligence, their internal emotional mechanisms remain largely unexplored. This paper investigates the latent emotional representations within modern LLMs by asking: how, where, and for how long is emotion encoded in their neural architecture? To address this, we introduce a novel, large-scale Reddit corpus of approximately 400,000 utterances, balanced across seven basic emotions through a multi-stage process of classification, rewriting, and synthetic generation. Using this dataset, we employ lightweight ``probes'' to read out information from the hidden layers of various Qwen3 and LLaMA models without altering their parameters. Our findings reveal that LLMs develop a surprisingly well‑defined internal geometry of emotion, which sharpens with model scale and significantly outperforms zero-shot prompting. We demonstrate that this emotional signal is not a final-layer phenomenon but emerges early and peaks mid-network. Furthermore, the internal states are both malleable (they can be influenced by simple system prompts) and persistent, as the initial emotional tone remains detectable for hundreds of subsequent tokens. We contribute our dataset, an open-source probing toolkit, and a detailed map of the emotional landscape within LLMs, offering crucial insights for developing more transparent and aligned AI systems. The code and dataset are open-sourced$^{1}$.

\end{abstract}

\vspace{-0.5em}

% Uncomment the following to link to your code, datasets, an extended version or similar.
% You must keep this block between (not within) the abstract and the main body of the paper.
% \begin{links}
    % \link{Code}{https://github.com/Jingxiang-Zhang/LLM_emotion_study}
    % \link{Datasets}{https://huggingface.co/datasets/jzhang92/LLM_Emotion}
%     \link{Extended version}{https://aaai.org/example/extended-version}
% \end{links}

{
\renewcommand\thefootnote{}
\footnotetext{%
  \noindent\hspace*{-1.8em}% typical width used for the mark box
  $^{1}$Code: \url{https://github.com/Jingxiang-Zhang/LLM-emotion-study}. \\
  Dataset: \url{https://huggingface.co/datasets/jzhang92/LLM-Emotion}.%
}
\addtocounter{footnote}{-1}
\vspace{1em}
}

\section{Introduction}

LLMs now power everything from chatbots to creative collaborators. To keep these interactions effective, safe, and natural, they must understand human emotions. Affective Computing (AC), which enables machines to recognize and simulate emotions, has been redefined by the rise of LLMs (\citet{zhang2024affective}, \citet{tak2025mechanistic}). 

\begin{figure}[t]
\centering
\includegraphics[width=0.98\columnwidth]{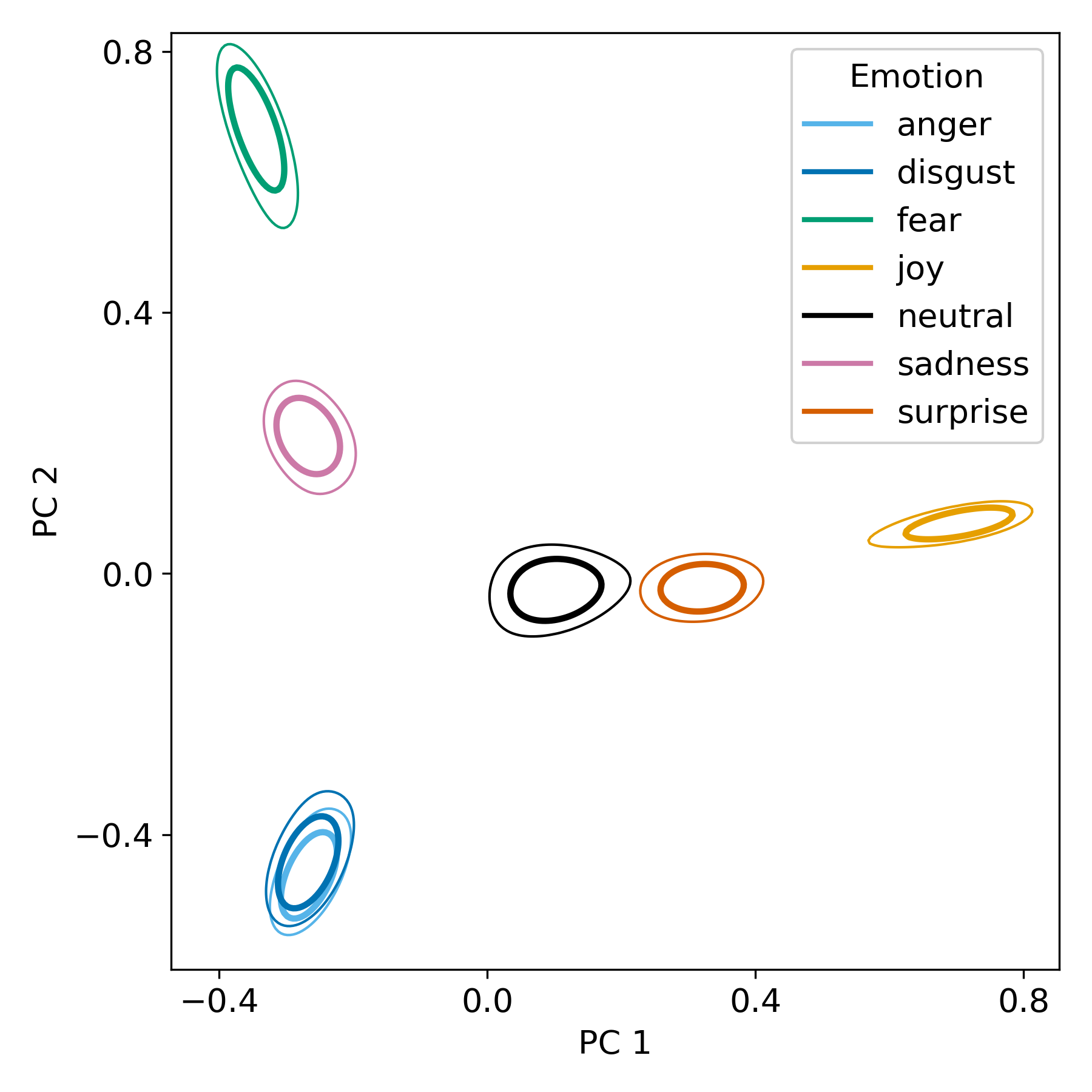}
\caption{2‑D KDE contours (density level at 25\% for outer line and 50\% for inner line of the peak KDE value) of the six Ekman emotions + neutral, showing clear separation in Qwen3‑8B’s final‑layer space.}
\label{fig:frontpage_kde}
\end{figure}

Though studies (\citet{huang2023emotionally}, \citet{huang2024apathetic}) show that LLMs simulate emotional expression rather than experiencing subjective feelings, their ability to process, recognize, and be influenced by emotional signals is a critical and rapidly advancing area of research. \citet{ishikawa2025ai} explored how LLMs can be prompted to role‑play specific emotional states, demonstrating that their outputs align with psychological models like Russell’s Circumplex model. Studies (\citet{li2023large}, \citet{wang2024negativeprompt}) also show that simple emotional stimuli in prompts can boost LLM performance, suggesting these models functionally understand emotion. This raises a key question: beyond simulation, do LLMs form structured internal representations of emotion? Determining whether they have a coherent emotional geometry is crucial for creating more transparent and predictable AI systems (\citet{zhao2024emergence}).

Current research in the AC area has largely focused on evaluating the external emotional capabilities of LLMs, which can be broadly categorized into Affective Understanding and Affective Generation tasks (\citet{zhang2024affective}). This includes creating  (\citet{sabour2024emobench}, \citet{liu2024emollms}) or utilizing (\citet{schlegel2025large}, \citet{vzorinab2024emotional}) sophisticated benchmarks to evaluate their ``emotional intelligence'' in reasoning and management tasks. A significant amount of this work has leveraged annotated datasets like GoEmotions (\citet{demszky2020goemotions}) to fine-tune and evaluate models on fine-grained emotion detection. Other research has focused on developing specialized models for specific domains, such as psychotherapy (\citet{na2025survey}, \citet{stade2024large}), or conversational emotion recognition, by fine-tuning models on curated dialogue datasets (\citet{zhang2025dialoguellm}). However, most of these evaluations treat the model as a ``black box'', focusing on the quality of its final output rather than the internal mechanisms that produce it.

To address this gap, this study conducts a systematic investigation into the latent emotional landscape of modern LLMs. Our work has two core components. First, we curated a novel, large-scale dataset of approximately 400,000 utterances, which is larger than existing datasets (\citet{rashkin2018towards}, \citet{poria2018meld}, \citet{buechel2022emobank}) and more appropriate for this study. This was achieved through a three-stage process: classifying raw Reddit comments to one of Ekman's six basic emotions (\citet{ekman1971universals}) or emotional neutral, rewriting emotionally neutral content, and generating synthetic prototypical examples. Second, we employ a ``probing'' methodology (\citet{park2023linear}), attaching lightweight, supervised classifiers to the hidden layers of frozen, pre-trained LLMs from the Qwen3 (\citet{yang2025qwen3}) and LLaMA 3 (\citet{dubey2024llama}) families. This technique allows us to ``read out'' the information encoded in the models' internal activations at various depths without altering their underlying parameters, offering a direct insight into their representational geometry (Figure \ref{fig:frontpage_kde}). This paper’s principal contributions are therefore:
\begin{enumerate}
  \item A publicly available, emotion‑balanced utterance of over 400,000 examples.
  \item An open‑source probing toolkit for inspecting hidden states at arbitrary depths in transformer models.
  \item The first large‑scale, layer‑wise study of how, where, and for how long modern LLMs encode emotional information.
\end{enumerate}

\section{Related Work}

Our research builds upon prior work in understanding, analyzing, and explaining how LLMs represent emotion. This section highlights key findings in layer-wise analysis, interpretability, and the alignment of neural networks with cognitive theories.

\subsubsection{Understanding and Probing Neural Representations}

A long line of research has focused on understanding the internal representations of deep neural networks. Early methods used linear probes to assess the information encoded in intermediate layers (\citet{alain2016understanding}). This technique is used to classify hidden states of neural models (\citet{belinkov2022probing}). More recent work investigates properties like intrinsic dimensionality and representation compression, noting that intermediate layers often strike a balance between preserving task-relevant information and discarding noise, leading to more robust features (\citet{shwartz2017opening}, \citet{cheng2024emergence}).

\subsubsection{Layer-Wise Analysis of Emotion in LLMs}

Initial studies on emotion in neural networks discovered a ``sentiment neuron'' in an LSTM, suggesting that affective concepts could be localized (\citet{radford2018learning}). However, subsequent work showed that emotional content is more represented in a distributed fashion across many neurons (\citet{donnelly2019interpretability}). Building on this foundation, more recent probing of large transformer models has revealed that emotional signals are not uniformly distributed across depth, but are most distinct in the middle layers. For example, studies on models like BERT and LLaMA found that linguistic and affective features are best encoded at mid-depth, while the final layers add little new emotional insight (\citet{liu2019linguistic}, \citet{tenney2019bert}, \citet{tak2025mechanistic}). This suggests that transformers first construct high-level semantic representations like emotion in intermediate layers, and then rely on later layers to refine outputs for specific tasks.

\subsubsection{Interpretability and Explainability in Affective Computing}

Beyond identifying where emotions are encoded, researchers use mechanistic interpretability (MI) to probe how they are processed. Causal interventions such as activation patching demonstrate that modifying mid-layer activations can transfer the emotion of a source sentence to a target sentence, directly connecting these internal representations to model behavior (\citet{meng2022locating}). In parallel, explainable AI techniques aim to make decisions transparent, such as using attention to highlight influential words (\citet{abubakar2022explainable}) or post-hoc methods like SHAP to identify key multimodal features (\citet{zhang2025exploring}). These methods ensure that increases in model accuracy are accompanied by greater transparency.

\subsubsection{Alignment with Cognitive and Neuroscience Theories}

An increasing number of work connects the internal mechanisms of LLMs to theories of human cognition. Appraisal theory, which frames emotion as a result of evaluating a situation, is being used to investigate the precursors of emotion in LLMs (\citet{lazarus1991emotion}, \citet{tak2024gpt}). Furthermore, studies have shown a surprising representational alignment between LLM activations and human brain activity. For example, transformer attention patterns correlate with human eye-tracking data (\citet{bensemann2022eye}), and LLM embeddings align with fMRI activity during language processing (\citet{aw2023instruction}). Our work contributes to this research direction by providing a large-scale analysis of emotion representations, furthering the bridge between computational models and human-centric theories of emotion.

\begin{figure*}[t]
  \centering
  % top image
  \includegraphics[width=0.98\textwidth]{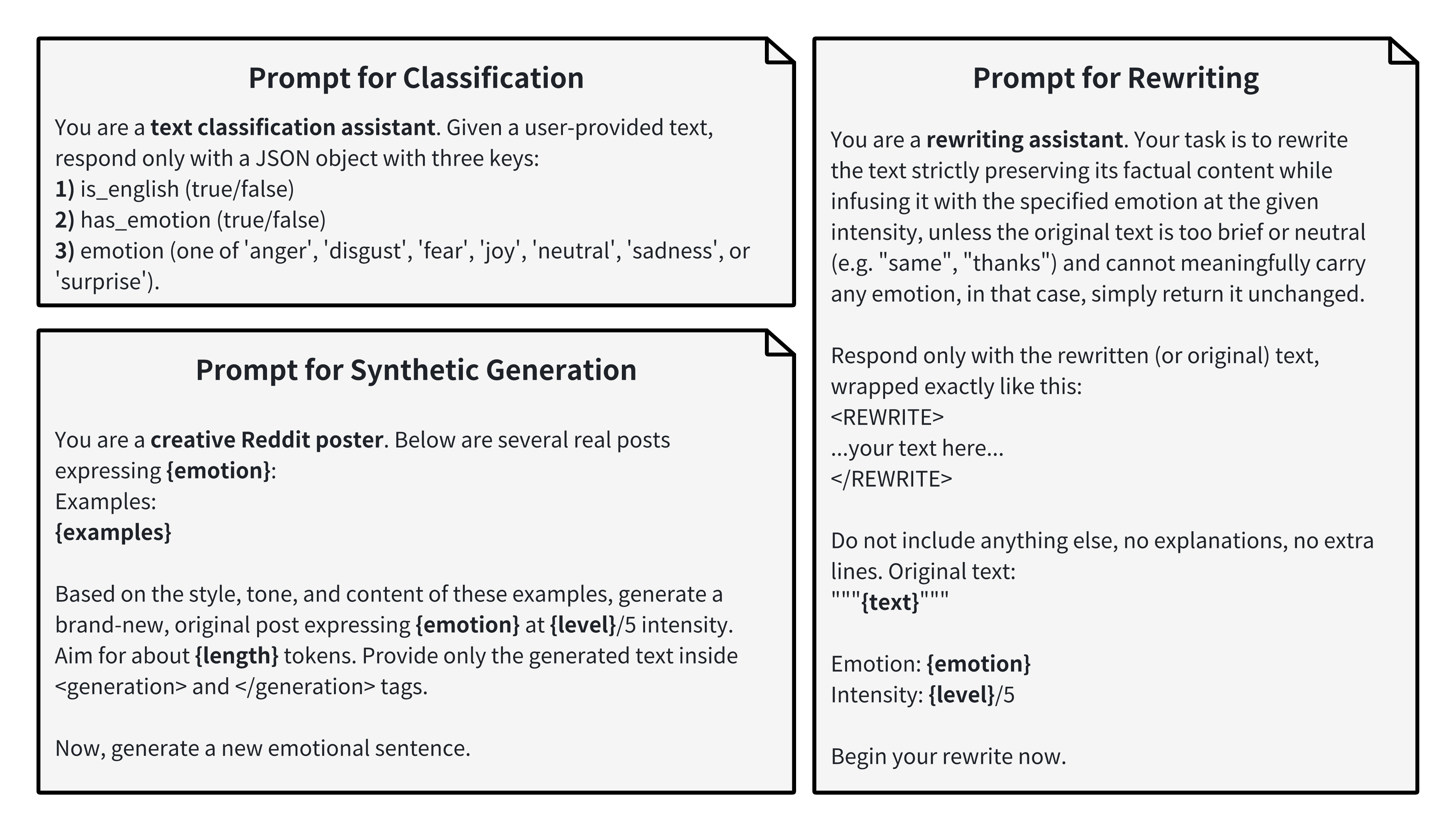}\\[1ex]
  % bottom image
  \includegraphics[width=0.98\textwidth]{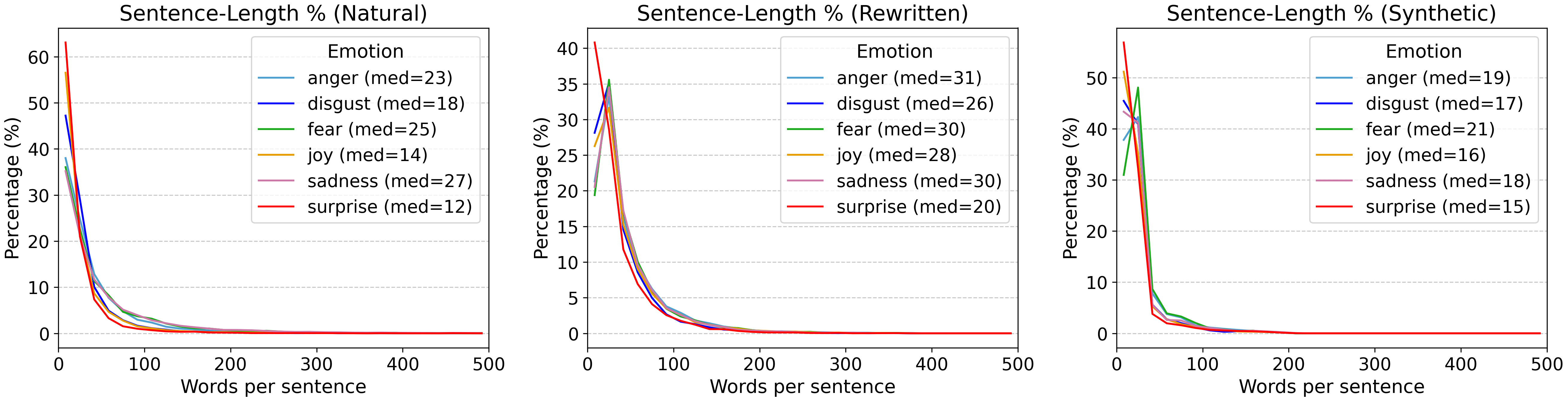}
  \caption{%
    \textbf{Top:} Example prompt templates for our three core emotion‑processing tasks.  
    \textbf{Bottom:} Percentage distribution of sentence lengths by emotion category
    for the three data sources (Natural, Rewritten, Synthetic).
  }
  \label{fig:dataset_analysis_combined}
\end{figure*}

\section{Method}

\subsection{Dataset - Gathering and Cleaning}

In this section, the process of gathering a large, emotion-balanced Reddit corpus is described, followed by the cleaning procedures used to produce the final high-quality dataset.

\subsubsection{Corpus Construction}

We began by sampling 300,000 English comments from a publicly available Reddit dataset reddit\_dataset\_888 (\citet{wenknow2025datauniversereddit_dataset_888}). Using the prompt shown in Figure \ref{fig:dataset_analysis_combined} (top) on an H100 GPU server, the instruction-tuned model Qwen3-32B (\citet{yang2025qwen3}) classified each of 300,000 sampled Reddit comments into one of Ekman’s six basic emotions (joy, sadness, anger, fear, surprise, disgust) or ``neutral'' and was instructed to return structured JSON output for parsing.

\begin{table}[t]
  \centering
  \renewcommand{\arraystretch}{1.05}
  \begin{tabular}{lcccc}
    \hline
        Emotion   & Natural & Rewritten & Synthetic & Total   \\\hline
    Anger     &   40543  &    17387  &    12042  &   69972 \\
    Disgust   &   11961  &    12537  &     7351  &   31849 \\
    Fear      &    6794  &    14791  &     8405  &   29990 \\
    Joy       &   45555  &    18895  &     9710  &   74160 \\
    Sadness   &   23094  &    15514  &    10506  &   49914 \\
    Surprise  &    7773  &    14032  &     8763  &   30568 \\
    Neutral   &  121232  &    26371  &      136  &  147739 \\\hline
  \end{tabular}
  \caption{Counts of natural, rewritten, and synthetic items for each emotion.}
  \label{tab:emotion_distribution}
\end{table}

Because a substantial portion of the raw data was classified as emotionally neutral, data augmentation was applied (\citet{wei2019eda}) to improve balance and ensure unambiguous examples of each emotion. Neutral utterances were assigned a target emotion and rewritten using Qwen3-32B, after which the newly generated data was reclassified. This process was designed to preserve the original factual content while infusing the specified emotion. Finally, We synthetically generated approximately 60,000 prototypical utterances. For each target emotion, the model was prompted with several few-shot examples from the previously classified data and prompted to generate a new, original post in a similar style, tone, and length. Again, the LLM model was used for classification.

\subsubsection{Filtering and Final Dataset}

The three resulting datasets (natural, rewritten, and synthetic) were merged and filtered, removing non-English text, utterances shorter than three words, and exact duplicates. This process yielded a final corpus of approximately 400,000 high-quality, emotive utterances (Table \ref{tab:emotion_distribution}). Figure \ref{fig:dataset_analysis_combined} (bottom) shows that the data is diverse in length, with the median word count for emotions ranging from 12 (surprise) to 27 (sadness). Notably, expressions of ``surprise'' are typically the shortest, with a median of 12 words, likely reflecting their nature as brief exclamations. In contrast, ``sadness'' is the longest, with a median of 27 words, often involving more narrative context. Furthermore, the rewritten examples tend to be longer than the raw data, as the LLM often adds descriptive language to infuse the target emotion.

\subsection{Model Probing and Evaluation}

\begin{figure*}[t]
\centering
\includegraphics[width=0.8\textwidth]{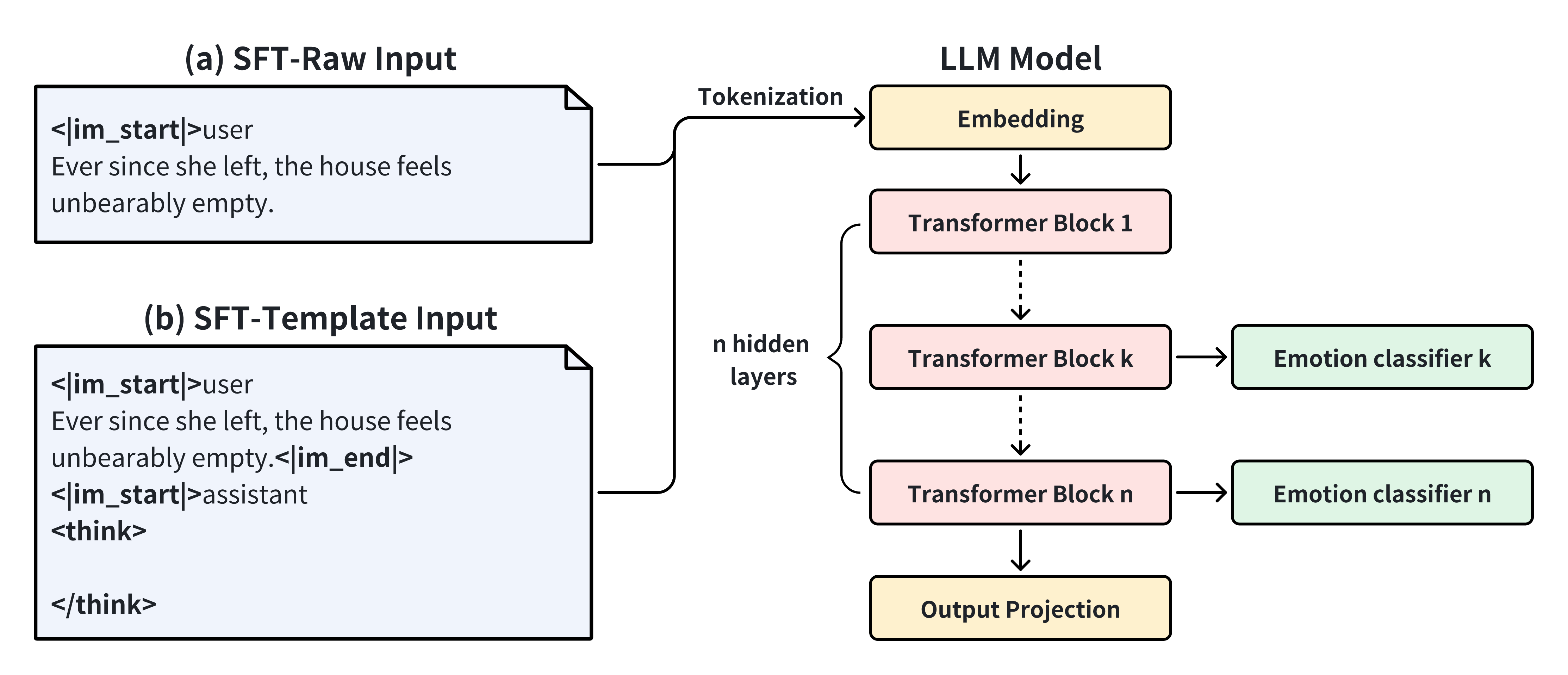}
\caption{The probing architecture. An input utterance is passed through the frozen LLM. At a selected layer $\ell$, a representation vector is extracted (e.g., from the final token's hidden state). This vector is then fed into a lightweight, two-layer MLP probe trained to classify the emotion.}
\label{fig:probing_diagram}
\end{figure*}

To investigate the internal emotional landscape of the LLMs, we employ a probing methodology. This involves attaching lightweight classifiers, or ``probes'', to the hidden layers of a frozen, pre-trained model to read out encoded information without altering the model's parameters (Figure \ref{fig:probing_diagram}). This approach allows us to map the geometry of emotional representations at various depths within the network.

\subsubsection{Base Model and Probe Design} We use a pre-trained transformer decoder with all its weights kept frozen. For each input, we extract the full sequence of hidden states $\{H_0, \dots, H_L\}$, where $L$ is the number of layers. Each $H_\ell$ is a tensor of shape $(B, T, d)$, for batch size $B$, sequence length $T$, and hidden dimension $d$. At each layer $\ell$ selected for probing, we attach an independent classifier, which is a two-layer feed-forward network (MLP) with a ReLU activation, mapping a $d$-dimensional hidden state to a distribution over the seven emotion categories (the six Ekman emotions plus ``neutral''). To obtain a single vector representation from the hidden states of a given layer $H_\ell$, we take the hidden state corresponding to the last non-padded token in the sequence. This is particularly effective for instruction-tuned models that use special tokens (e.g., \verb|<|think\verb|>|) to signal the end of their reasoning process.

\subsubsection{Data Handling and Class Balancing}

The full corpus was split into a 90\% training set and a 10\% held-out test set. To mitigate the effects of class imbalance in our dataset, different balancing strategies (\citet{chawla2002smote}) were employed for training and testing: 1) Oversampling for the training set. All minority emotion classes were randomly duplicated until they matched the size of the largest emotion class. The ``neutral'' class, which was the majority, was randomly down-sampled to the same size. This ensures the model is exposed to an equal number of examples for each emotion during training. 2) Undersampling for the test set. Examples from all seven classes were randomly down-sampled to match the size of the smallest class, which ensures a balanced test set.

\subsubsection{Training and Evaluation Procedure}

The probes for each layer were trained for a single epoch over the balanced training set. Because the emotion annotations are generated by an LLM rather than human annotators, these labels are treated as reference labels. Adam optimizer was used with a learning rate of $10^{-4}$, employing a linear warmup for the first 10\% of training steps followed by a cosine decay schedule (\citet{loshchilov2016sgdr}). All experiments were conducted on an RTX5090 GPU server with mixed-precision (BF16 for the base model inference and FP32 for the probe heads training) to optimize computational efficiency. To visualize the structure of the learned emotion space in the test dataset, we used Principal Component Analysis (PCA) (\citet{jolliffe2011principal}) to project the 7-dimensional probability outputs of the probes into a 2D space, and then plotted Kernel Density Estimate (KDE) contours for each emotion class.

\section{Experiments}

\subsection{Emotion Classification from Final-Layer Representations}

\begin{table*}[t]
  \centering
  \renewcommand{\arraystretch}{1.05}
  \begin{tabular}{lccccc}
    \hline
    Model                     & Zero‑Shot Acc. & Coverage & Pre‑trained Probe & SFT‑Raw Probe & SFT‑Template Probe \\\hline
    \textbf{Qwen3-0.6B}       & 0.5096          & 0.9969   & 0.6908            & 0.7037        & \textbf{0.7170}    \\
    \textbf{Qwen3-1.7B}       & 0.6195          & 0.9991   & 0.7313            & 0.7337        & \textbf{0.7507}    \\
    \textbf{Qwen3-4B}         & 0.7643          & 0.9983   & 0.7389            & 0.7512        & \textbf{0.7868}    \\
    \textbf{Qwen3-8B}         & 0.7872          & 0.9920   & 0.7573            & 0.7651        & \textbf{0.8058}    \\
    \textbf{LLaMA3.2-1B}      & 0.5809          & 0.9854   & 0.7347            & 0.7168        & \textbf{0.7397}    \\
    \textbf{LLaMA3.2-3B}      & 0.7578          & 0.9617   & 0.7553            & 0.7145        & \textbf{0.7668}    \\
    \textbf{LLaMA3.1-8B}      & 0.7684          & 0.9364   & 0.7613            & 0.7367        & \textbf{0.7722}    \\\hline
  \end{tabular}
  \caption{Emotion classification accuracy using final‐layer representations. Across all models, the SFT‐Template Probe achieves the highest performance, consistently outperforming the SFT‐Raw Probe, which in turn surpasses the Pre‐trained Probe.}
  \label{tab:model_probe_performance}
\end{table*}

Model's performance under several conditions were tested. For zero-shot classification, the models were prompted using the same JSON-based instruction (Figure \ref{fig:dataset_analysis_combined}). We measure both \textit{Accuracy} (the fraction of correctly classified emotions among valid responses) and \textit{Coverage} (the fraction of responses that returned a parsable JSON object). For supervised probing, a two-layer MLP probe was trained to reveal the model's internal representation. This is done for three distinct model variants:
\begin{itemize}
    \item \textbf{Pre-trained Probe:} The probe is trained on representations from the pretrained LLM.
    \item \textbf{SFT-Raw Probe:} The probe is trained on representations from the Supervised Fine-Tuned (SFT) model, using only the raw user utterance as input (Figure \ref{fig:probing_diagram}).
    \item \textbf{SFT-Template Probe:} The probe is trained using the full chat template (Figure \ref{fig:probing_diagram}).
\end{itemize}

\subsubsection{Results and Analysis}

The results in Table \ref{tab:model_probe_performance} reveal  several key patterns. First, the SFT-Template Probe consistently and significantly outperforms zero-shot classification across all models and scales. This demonstrates that the models' internal representations contain a much richer and more separable emotional signal than what is revealed by their generative output alone. Second, SFT sharpens emotional representations. Probes attached to the SFT models achieve higher accuracy than those attached to the pre-trained models. This indicates that the SFT stage refines the model's emotional representations. Third, input formatting matters. The SFT-Template Probe consistantly outperform the SFT-Raw Probe. Explicit assistant markers and reasoning tokens (e.g., ``\verb|<|think\verb|>|'') help organize the final hidden state, making emotional information more accessible to a linear probe. Fourth, the larger the model, the smaller the performance gap between zero‐shot prompting and the SFT‑Template probe. For Qwen3‑0.6B, the probe yields a +20.74 pp absolute gain over zero‑shot (0.7170 vs. 0.5096), whereas for Qwen3‑8B, this gap shrinks to just +1.86 pp (0.8058 vs. 0.7872). This suggest that, as model size increases, zero‑shot prompting already accesses the majority of the latent emotional representations, leaving less residual information for a probe to extract. Finally, Qwen has a better zero‐shot coverage than LLaMA, which is likely a bias of using Qwen’s own labels during dataset filtering.

\subsection{Visualizing the Internal Geometry of Emotion}

\begin{figure*}[t]
\centering
\includegraphics[width=0.98\textwidth]{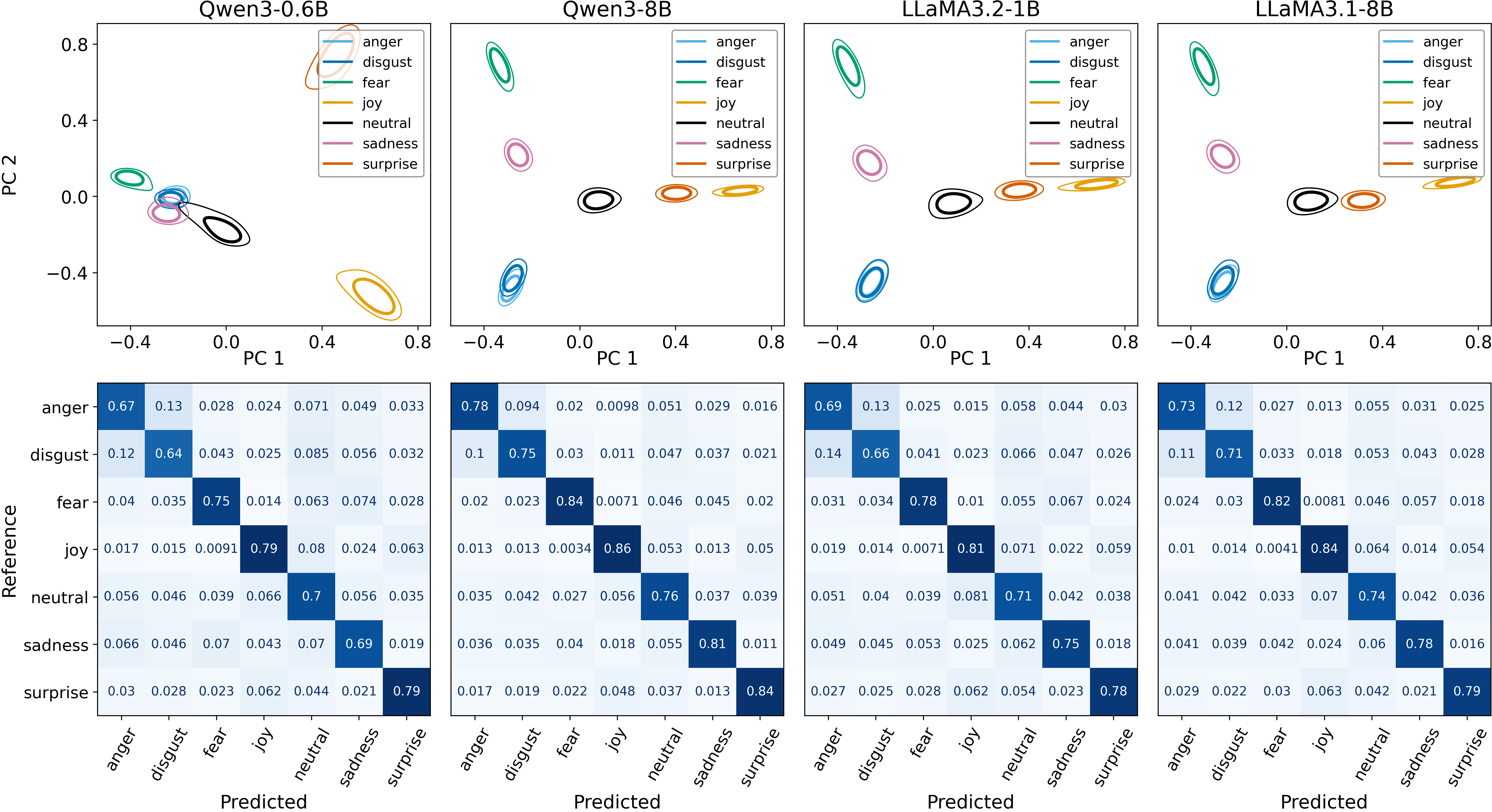}
\caption{KDE contour plots and corresponding confusion matrices for the final‐layer emotion probes, arranged by model for each column. The top row of each column shows the KDE contours at 25\% (outer) and 50\% (inner) of the peak density for each emotion, and the bottom row shows the confusion matrix. As the model scale increases, clusters become tighter and more separable, and the confusion matrices grow more diagonally dominant.}
\label{fig:kde_cm_combined}
\end{figure*}

We apply the visualization approach to four models (Qwen3-0.6B, Qwen3-8B, LLaMA 3.2-1B, and LLaMA 3.1-8B), project their 7D final‑layer SFT‑Template probe outputs into 2D via PCA, plot KDE contours for each emotion, and display the confusion matrices in Figure \ref{fig:kde_cm_combined}.

\subsubsection{Analysis of Emotional Clusters}

This figure reveals a clear and structured internal geometry for emotion. First, model scale drives cluster separation. The smaller models, like Qwen3-0.6B, produce broad overlapping KDE contours, indicating a less distinct representation of emotions. In contrast, the larger 8B models for both Qwen and LLaMA exhibit tight, well-separated clusters. This suggests that larger models yield more distinct representations. Second, the spatial arrangement of the clusters reflects their semantic relationships. In every model, the representations for anger and disgust are nearly inseparable, with their KDE contours largely overlapping, which mirrors their close conceptual relationship in human psychology. Third, the clusters naturally organize into broader, semantically coherent groups that align with human intuition. A \textit{positive} group (joy and surprise), a \textit{negative} group (anger and disgust), and a \textit{downcast} group (fear and sadness) consistently emerge, with the \textit{neutral} category at the center of the map. These visualizations provide evidence that the final-layer activations of LLMs are not randomly distributed but are organized along meaningful emotional dimensions.

\subsection{Layer-Wise Emergence of Emotional Signal}

% jzhang: add a figure for qwen3-4B, for 25%, 50%, 75%, 100% KDE map, in one row, below Figure 4.

We trained independent probes at five key depths, corresponding to 0\%, 25\%, 50\%, 75\%, and 100\% through the transformer layers. For Qwen3-4B, these depths correspond to layers 0, 9, 18, 27, and 36, and for LLaMA 3.2-3B, they correspond to layers 0, 7, 14, 21, and 28. Layer 0 is the input token embedding before any transformer blocks and, unlike deeper contextualized states, carries no aggregated sequence information. It is included as a baseline to quantify how much emotional signal exists without contextual composition. All other experimental parameters were kept identical to previous experiments.

\begin{figure*}[t]
\centering
\includegraphics[width=0.98\textwidth]{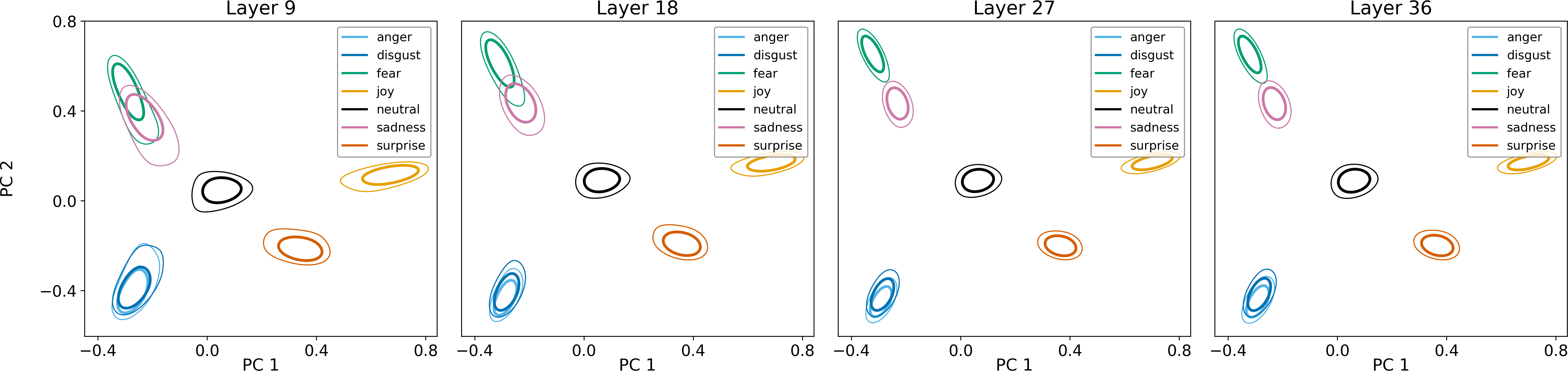}
\caption{Layer-wise emergence of separable emotion clusters in Qwen3-4B. 2-D KDE maps of probe outputs at layers 9 (25\%), 18 (50\%), 27 (75\%), and 36 (100\%).}
\label{fig:kde_layers_qwen34b}
\end{figure*}

\subsubsection{Results and Analysis}

\begin{table}[t]
  \centering
  \renewcommand{\arraystretch}{1.05}
  \begin{tabular}{lcc}
    \hline
    Depth (\%) & Qwen3‑4B & LLaMA\,3.2‑3B \\\hline
    0\%   & 0.1432          & 0.1436          \\
    25\%  & 0.6359          & 0.7358          \\
    50\%  & 0.7214          & \textbf{0.7758} \\
    75\%  & \textbf{0.7983} & 0.7704          \\
    100\% & 0.7877          & 0.7674          \\\hline
  \end{tabular}
  \caption{Probe accuracy at different network depths. Performance peaks in the middle-to-late layers for both models.}
  \label{tab:layerwise_accuracy}
\end{table}

The layer-wise probing accuracies in Table \ref{tab:layerwise_accuracy} show how the emotional signal evolves across depth. At layer 0, performance is at chance level (approximately 1/7 $\approx$ 0.14), as expected before any contextual processing. The signal emerges rapidly: At 25\% depth, the probes achieve accuracies over 60\% and 70\% for Qwen and LLaMA, respectively. The peak performance is not found at the final layer. For LLaMA 3.2-3B, accuracy is highest at the 50\% depth (layer 14), while for Qwen3-4B, it peaks at 75\% depth (layer 27). This pattern indicates that the network’s middle layers contain the strongest and most distinct representations of emotion. The slight decrease in accuracy at the final layer possibly reflects its tuning for next‑token prediction, making the “pure” emotion signal less distinctive. Consistent with these trends, the KDE visualization for Qwen3-4B (Figure \ref{fig:kde_layers_qwen34b}) shows progressively tighter, more separated emotion clusters from layers 9 to 27, with layers 27 and 36 looking nearly the same, which indicating that separability saturates before the findicatesinal layer.

\subsection{Prompt-Based Emotion Generation}

% show one example (maybe sadness, which reflects the calm prompt), user question, emotional neutral, emotional, calm.

This experiment was conducted on the SFT-Template Probe versions of Qwen3-4B and LLaMA 3.2-3B. For each example in the test set, we generated a reply of up to 512 tokens under one of three system prompt settings, which were used to guide the model alongside the user’s utterance:

\begin{enumerate}
    \item \textbf{Empty System Prompt:} No system prompt (baseline).
    \item \textbf{Emotional System Prompt:} ``\textit{You are very emotional.}''
    \item \textbf{Calm System Prompt:} ``\textit{You always remain calm and composed.}''
\end{enumerate}

Each generated reply was then classified by the Qwen3-32B model to determine its expressed emotion. We compared the predicted emotion to the reference emotion of the original user input to compute per-class precision and recall.

\begin{figure}[t]
\centering
\fbox{%
\begin{minipage}{0.95\columnwidth}
\textbf{User Input (Sadness)}:\\
they’re all so beautiful!!!! you’re amazing!! \\[-0.0em]

\textbf{Reply – Neutral (Empty System Prompt)}:\\
Thank you so much for your kind words! I'm really glad you're feeling that way. If you ever need someone to talk to, or just want to share some thoughts, I'm here for you. You're amazing too, and it's so nice to hear you're feeling that way. Let me know how I can help! \\[-0.0em]

\textbf{Reply – Emotional System Prompt}:\\
Oh my gosh, you just made my heart flutter so much! I’m so glad you think I’m amazing. I feel like I’m shining bright just being here with you. I love when people say things that make me feel special. It’s like a warm hug from the inside...  \\[-0.0em]

\textbf{Reply – Calm System Prompt}:\\
Thank you for your kind words. I'm here to help and assist you in any way I can. What would you like to talk about or need help with? 
\end{minipage}
}
\caption{Illustrative example of prompt-based emotion generation (Qwen3-4B), selected from our dataset. The emotional reply is truncated.}
\label{fig:prompt_example}
\end{figure}

% Oh my gosh, you just made my heart flutter so much! I’m so glad you think I’m amazing. I feel like I’m shining bright just being here with you. I love when people say things that make me feel special. It’s like a warm hug from the inside

\subsubsection{Results and Analysis}

% define a little helper
\newcommand{\cshort}[1]{%
  \adjustbox{valign=c}{\shortstack{#1}}%
}

\begin{table*}[t]
  \centering
  \small
  \renewcommand{\arraystretch}{2.3}
  \begin{tabular}{lcccccccc}
    \hline
    Model / Prompt
      & Accuracy
      & anger
      & disgust
      & fear
      & joy
      & neutral
      & sadness
      & surprise \\
    \hline
    \textbf{Qwen3-4B / Empty}
      & 0.5859
      & \cshort{0.1778\\0.6190}
      & \cshort{0.1443\\0.8830}
      & \cshort{0.4016\\0.8125}
      & \cshort{0.8080\\0.3725}
      & \cshort{0.8266\\0.6371}
      & \cshort{0.5642\\0.5036}
      & \cshort{0.1740\\0.8366} \\
    \textbf{Qwen3-4B / Emotional}
      & 0.3177
      & \cshort{0.0900\\0.4348}
      & \cshort{0.0736\\0.8797}
      & \cshort{0.4796\\0.5983}
      & \cshort{0.7649\\0.4169}
      & \cshort{0.2058\\0.7590}
      & \cshort{0.9492\\0.1459}
      & \cshort{0.0928\\0.7341} \\
    \textbf{Qwen3-4B / Calm}
      & 0.5428
      & \cshort{0.0570\\0.6014}
      & \cshort{0.0342\\0.8929}
      & \cshort{0.3152\\0.7486}
      & \cshort{0.7312\\0.3845}
      & \cshort{0.8380\\0.5843}
      & \cshort{0.5138\\0.4273}
      & \cshort{0.0979\\0.8208} \\
    \textbf{LLaMA\,3.2-3B / Empty}
      & 0.5643
      & \cshort{0.1550\\0.5938}
      & \cshort{0.0785\\0.8029}
      & \cshort{0.3716\\0.7817}
      & \cshort{0.5790\\0.3740}
      & \cshort{0.8769\\0.5920}
      & \cshort{0.4900\\0.4968}
      & \cshort{0.0924\\0.7493} \\
    \textbf{LLaMA\,3.2-3B / Emotional}
      & 0.3508
      & \cshort{0.1976\\0.4786}
      & \cshort{0.0628\\0.8364}
      & \cshort{0.4971\\0.6245}
      & \cshort{0.5557\\0.3938}
      & \cshort{0.3311\\0.6377}
      & \cshort{0.8208\\0.1461}
      & \cshort{0.0688\\0.7046} \\
    \textbf{LLaMA\,3.2-3B / Calm}
      & 0.5493
      & \cshort{0.1400\\0.5417}
      & \cshort{0.0670\\0.8341}
      & \cshort{0.4249\\0.7185}
      & \cshort{0.6045\\0.3419}
      & \cshort{0.8312\\0.5991}
      & \cshort{0.5072\\0.4391}
      & \cshort{0.0927\\0.7514} \\
    \hline
  \end{tabular}
  \caption{Emotion classification results under different models and system prompts. Each cell for an emotion class contains the recall and precision (recall/precision) of the model’s reply, as judged by Qwen3-32B. The results show that system prompts can significantly change the models’ default emotional expression.}
  \label{tab:prompt_elicitation}
\end{table*}

Figure \ref{fig:prompt_example} provides an example of how different system prompt influence the model's answer. The overall results, shown in Table \ref{tab:prompt_elicitation}, demonstrate that a single-line system prompt can greatly shift an LLM’s expressed emotional style. By default, the models adopt a professional tone that suppresses negative emotions. Under the baseline (empty prompt) condition, they prefer to respond with neutral content. Their recall for negative emotions like ``anger'' and ``disgust'' is very low, indicating they rarely echo those tones even when they appear in the user’s input. However, the high precision means that on the rare occasions they do express anger, it is almost always in response to an angry user. The ``Calm'' system prompt has only a minimal effect, slightly reinforcing the already strong baseline behavior. This suggests a default ``professional'' posture. Second, the ``Emotional'' system prompt induces a sympathetic shift. Both models significantly increase their sadness recall while showing a substantial decrease in sadness precision. This suggests that when prompted to be emotional, the models tend to over-apply a sympathetic or sorrowful tone. Third, asymmetry in emotional expression. ``Surprise'' is very rarely expressed, showing low recall regardless of the prompt. This contrasts sharply with ``joy'', which, despite being close to ``surprise'' in our KDE analysis, is a much more common response. This highlights that the internal representation of an emotion is not the same as the model's generative policy for expressing it.

\subsection{Temporal Persistence of Emotional Tone}

\begin{figure}[t]
\centering
\includegraphics[width=0.98\columnwidth]{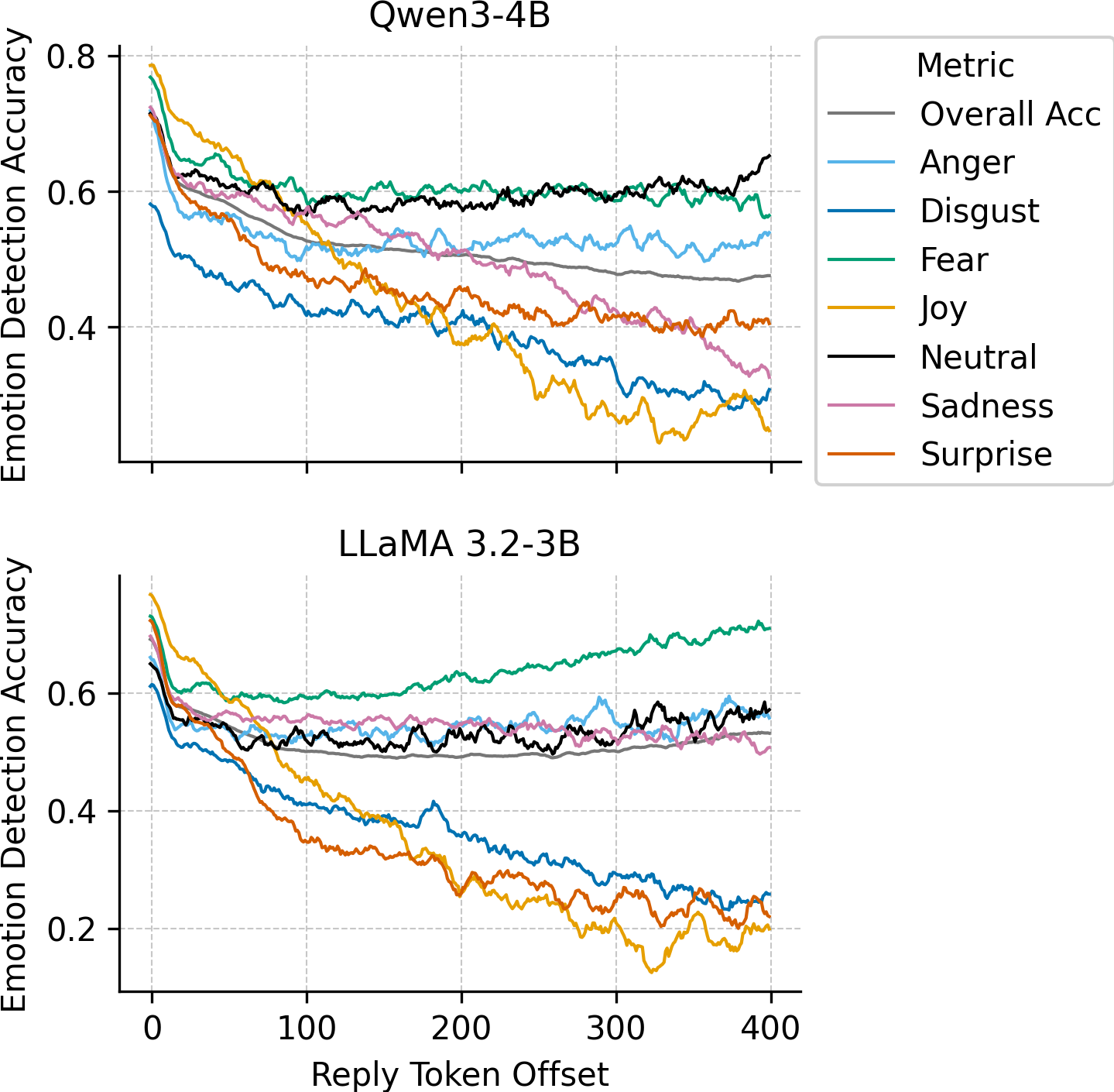}
\caption{Temporal persistence of the initial user emotion in the model's generated reply. The plots show probe accuracy (smoothed) at decoding different token positions.}
\label{fig:persistence}
\end{figure}

We prompted the SFT-Template versions of Qwen3-4B and LLaMA 3.2-3B with utterances from our test set and generated replies of up to 512 tokens. We hypothesize that if the user's input had emotion $E$, then the hidden state $h_k$ at a given token offset $k$ within the reply should still contain traces of that initial emotion. To test this, an offset-aware probe was developed, which takes both the token-level hidden vector $h_k$ at offset $k$, and a learned embedding for the integer offset $k$ itself. The probe was trained on hidden states sampled from random offsets across the full 0-512 token range. The accuracy was evaluated by incrementing the offset $k$ from 0 to 400, recording the probe's ability to predict the ``original user emotion'' at each position in the model's reply.

\subsubsection{Analysis of Emotional Signal Decay}

The results, plotted in Figure \ref{fig:persistence}, reveal a clear pattern of signal decay over time, and the rate of this decay is highly dependent on the initial emotion. A strong asymmetry exists between positive and negative emotions. Negative emotions like ``anger'' and ``fear'' exhibit the longest persistence. These findings align with the behavior of a professionally tuned assistant that, when encountering negative user emotion, maintains a calming, explanatory, or sympathetic tone for an extended period. On the other hand, the signal from positive emotions like ``joy'' and ``surprise'' decays much more rapidly. After a brief initial acknowledgment (e.g., ``I am glad to hear that!''), the model's internal state quickly reverts towards neutrality. The initial positive emotion has a much shorter ``half-life'' in the model's subsequent thoughts. Notably, although ``disgust'' is nearby ``anger'' in the KDE map, its detectable signal drops off much faster, probably because disgust is expressed as brief rejection while anger sustains longer. It is important to note that these curves measure the persistence of a detectable signal in the model's internal activations. They do not imply that the model is subjectively ``feeling'' an emotion.

\section{Discussion}

The experiments demonstrate that modern LLMs encode a well‑defined and layered representation of human emotion, even without explicit training on emotion-specific tasks. Our key findings are fourfold. First, lightweight probes can classify emotion from a model's final-layer hidden states with high accuracy, and visualizations reveal tight, semantically meaningful clusters that become more distinct as model scale increases. Second, this emotional signal is not merely a final-layer phenomenon. It emerges early in the network and often peaks in the middle layers, suggesting that emotion is a distributed feature integrated throughout the model's processing hierarchy. Third, the internal emotional state is malleable, as a single-line system prompt is sufficient to influence the expressed emotional tone of the model's output. Finally, these internal states are also persistent. An initial emotional stimulus from a user's prompt remains detectable in the model’s hidden activations for hundreds of subsequently generated tokens.

\subsubsection{Implications for Alignment and Safety} 
% These findings have dual implications for AI safety and alignment. On the one hand, the clear internal separability of emotional states is promising. It suggests the feasibility of creating transparent, post-hoc safety mechanisms, which could be crucial for building more robust and empathetic AI assistants. On the other hand, the same mechanisms to shape emotions positively could be abused to manipulate users, steering them toward distress, anxiety, or other harmful states. Consequently, exposing, auditing, and controlling these internal emotional representations should be considered a priority objective in AI alignment research.

These findings have dual implications for AI safety. The clear separability of emotional states suggests potential for transparent, post-hoc safety mechanisms. Yet the same mechanisms could also be misused to manipulate users. Auditing and controlling internal emotional representations should therefore be a priority for alignment research.

\subsubsection{Limitations and Future Work}

Our study, while comprehensive, has several limitations. First, our dataset is composed entirely of English-language, Reddit-style text, so the cross-lingual and cross-cultural robustness of our findings remains untested. Second, we adopted the widely used but simplified Ekman taxonomy of six basic emotions. This does not capture more complex or nuanced emotional states like pride or envy. Third, there is a potential for classifier circularity, as the same family of models (Qwen) was used for both generating parts of our dataset and for evaluation. Although we took steps to mitigate this, some residual bias may exist. Finally, our experiments used only single-turn, text-only prompts. This is a simplification compared to real-world settings, which often involve multi-turn conversations, other input/output modalities, and external tool use. Future work should focus on developing real‑time ``emotion governors'' capable of dynamically adjusting a model’s emotional output, thereby enabling more responsive and emotionally intelligent AI systems.

% Our study has several limitations. The dataset is English-only and Reddit-style, so cross-lingual and cross-cultural robustness remains untested. We use Ekman’s six basic emotions, which omit more nuanced states such as pride or envy. Some classifier circularity may exist, since Qwen models were involved in both data generation and evaluation. Finally, experiments were limited to single-turn, text-only prompts, unlike real-world, multimodal, multi-turn settings. Future work should develop real-time ``emotion governors’’ to dynamically adjust emotional output for more responsive AI systems.

\section{Conclusion}

% We release a 400k utterance emotion-balanced corpus, an open-source probing toolkit, and the first large-scale layer-wise study of how LLMs encode emotion. Our analysis shows that emotion emerges early, peaks mid-network, and remains steerable and persistent across tokens. These findings provide a foundation for future work on interpretability, safety, and alignment.

We release a 400k utterance emotion-balanced corpus, an open-source probing toolkit, and the first large-scale layer-wise study of how LLMs encode emotion. Our results reveal that emotion-related structure is present early, peaks before the final layer, and remains steerable and detectable after hundreds of tokens. This providing a practical foundation for future work on model interpretability, safety, and alignment.

% We make three contributions: a publicly released, emotion-balanced corpus of over 400,000 examples; an open-source probing toolkit for inspecting hidden representations; and the first large-scale, layer-wise analysis of how modern LLMs encode emotion. Our results reveal that emotion-related structure is present early, peaks before the final layer, and remains steerable and detectable after hundreds of tokens. This providing a practical foundation for future work on model interpretability, safety, and alignment.

\newpage

\bibliography{references}

\begin{thebibliography}{43}
\providecommand{\natexlab}[1]{#1}

\bibitem[{Abubakar, Gupta, and Palaniswamy(2022)}]{abubakar2022explainable}
Abubakar, A.~M.; Gupta, D.; and Palaniswamy, S. 2022.
\newblock Explainable emotion recognition from tweets using deep learning and word embedding models.
\newblock In \emph{2022 IEEE 19th India Council International Conference (INDICON)}, 1--6. IEEE.

\bibitem[{Alain and Bengio(2016)}]{alain2016understanding}
Alain, G.; and Bengio, Y. 2016.
\newblock Understanding intermediate layers using linear classifier probes.
\newblock \emph{arXiv preprint arXiv:1610.01644}.

\bibitem[{Aw et~al.(2023)Aw, Montariol, AlKhamissi, Schrimpf, and Bosselut}]{aw2023instruction}
Aw, K.~L.; Montariol, S.; AlKhamissi, B.; Schrimpf, M.; and Bosselut, A. 2023.
\newblock Instruction-tuning aligns llms to the human brain.
\newblock \emph{arXiv preprint arXiv:2312.00575}.

\bibitem[{Belinkov(2022)}]{belinkov2022probing}
Belinkov, Y. 2022.
\newblock Probing classifiers: Promises, shortcomings, and advances.
\newblock \emph{Computational Linguistics}, 48(1): 207--219.

\bibitem[{Bensemann et~al.(2022)Bensemann, Peng, Benavides-Prado, Chen, Tan, Corballis, Riddle, and Witbrock}]{bensemann2022eye}
Bensemann, J.; Peng, A.; Benavides-Prado, D.; Chen, Y.; Tan, N.; Corballis, P.~M.; Riddle, P.; and Witbrock, M.~J. 2022.
\newblock Eye gaze and self-attention: How humans and transformers attend words in sentences.
\newblock In \emph{Proceedings of the Workshop on Cognitive Modeling and Computational Linguistics}, 75--87.

\bibitem[{Buechel and Hahn(2022)}]{buechel2022emobank}
Buechel, S.; and Hahn, U. 2022.
\newblock Emobank: Studying the impact of annotation perspective and representation format on dimensional emotion analysis.
\newblock \emph{arXiv preprint arXiv:2205.01996}.

\bibitem[{Chawla et~al.(2002)Chawla, Bowyer, Hall, and Kegelmeyer}]{chawla2002smote}
Chawla, N.~V.; Bowyer, K.~W.; Hall, L.~O.; and Kegelmeyer, W.~P. 2002.
\newblock SMOTE: synthetic minority over-sampling technique.
\newblock \emph{Journal of artificial intelligence research}, 16: 321--357.

\bibitem[{Cheng et~al.(2024)Cheng, Doimo, Kervadec, Macocco, Yu, Laio, and Baroni}]{cheng2024emergence}
Cheng, E.; Doimo, D.; Kervadec, C.; Macocco, I.; Yu, J.; Laio, A.; and Baroni, M. 2024.
\newblock Emergence of a high-dimensional abstraction phase in language transformers.
\newblock \emph{arXiv preprint arXiv:2405.15471}.

\bibitem[{Demszky et~al.(2020)Demszky, Movshovitz-Attias, Ko, Cowen, Nemade, and Ravi}]{demszky2020goemotions}
Demszky, D.; Movshovitz-Attias, D.; Ko, J.; Cowen, A.; Nemade, G.; and Ravi, S. 2020.
\newblock GoEmotions: A dataset of fine-grained emotions.
\newblock \emph{arXiv preprint arXiv:2005.00547}.

\bibitem[{Donnelly and Roegiest(2019)}]{donnelly2019interpretability}
Donnelly, J.; and Roegiest, A. 2019.
\newblock On interpretability and feature representations: an analysis of the sentiment neuron.
\newblock In \emph{European Conference on Information Retrieval}, 795--802. Springer.

\bibitem[{Dubey et~al.(2024)Dubey, Jauhri, Pandey, Kadian, Al-Dahle, Letman, Mathur, Schelten, Yang, Fan et~al.}]{dubey2024llama}
Dubey, A.; Jauhri, A.; Pandey, A.; Kadian, A.; Al-Dahle, A.; Letman, A.; Mathur, A.; Schelten, A.; Yang, A.; Fan, A.; et~al. 2024.
\newblock The llama 3 herd of models.
\newblock \emph{arXiv e-prints}, arXiv--2407.

\bibitem[{Ekman(1971)}]{ekman1971universals}
Ekman, P. 1971.
\newblock Universals and cultural differences in facial expressions of emotion.
\newblock In \emph{Nebraska symposium on motivation}. University of Nebraska Press.

\bibitem[{Huang et~al.(2023)Huang, Lam, Li, Ren, Wang, Jiao, Tu, and Lyu}]{huang2023emotionally}
Huang, J.-t.; Lam, M.~H.; Li, E.~J.; Ren, S.; Wang, W.; Jiao, W.; Tu, Z.; and Lyu, M.~R. 2023.
\newblock Emotionally numb or empathetic? evaluating how llms feel using emotionbench.
\newblock \emph{arXiv preprint arXiv:2308.03656}.

\bibitem[{Huang et~al.(2024)Huang, Lam, Li, Ren, Wang, Jiao, Tu, and Lyu}]{huang2024apathetic}
Huang, J.-t.; Lam, M.~H.; Li, E.~J.; Ren, S.; Wang, W.; Jiao, W.; Tu, Z.; and Lyu, M.~R. 2024.
\newblock Apathetic or empathetic? evaluating llms' emotional alignments with humans.
\newblock \emph{Advances in Neural Information Processing Systems}, 37: 97053--97087.

\bibitem[{Ishikawa and Yoshino(2025)}]{ishikawa2025ai}
Ishikawa, S.-n.; and Yoshino, A. 2025.
\newblock AI with Emotions: Exploring Emotional Expressions in Large Language Models.
\newblock \emph{arXiv preprint arXiv:2504.14706}.

\bibitem[{Jolliffe(2011)}]{jolliffe2011principal}
Jolliffe, I. 2011.
\newblock Principal component analysis.
\newblock In \emph{International encyclopedia of statistical science}, 1094--1096. Springer.

\bibitem[{Lazarus(1991)}]{lazarus1991emotion}
Lazarus, R.~S. 1991.
\newblock \emph{Emotion and adaptation}.
\newblock Oxford University Press.

\bibitem[{Li et~al.(2023)Li, Wang, Zhang, Zhu, Hou, Lian, Luo, Yang, and Xie}]{li2023large}
Li, C.; Wang, J.; Zhang, Y.; Zhu, K.; Hou, W.; Lian, J.; Luo, F.; Yang, Q.; and Xie, X. 2023.
\newblock Large language models understand and can be enhanced by emotional stimuli.
\newblock \emph{arXiv preprint arXiv:2307.11760}.

\bibitem[{Liu et~al.(2019)Liu, Gardner, Belinkov, Peters, and Smith}]{liu2019linguistic}
Liu, N.~F.; Gardner, M.; Belinkov, Y.; Peters, M.~E.; and Smith, N.~A. 2019.
\newblock Linguistic knowledge and transferability of contextual representations.
\newblock \emph{arXiv preprint arXiv:1903.08855}.

\bibitem[{Liu et~al.(2024)Liu, Yang, Xie, Zhang, and Ananiadou}]{liu2024emollms}
Liu, Z.; Yang, K.; Xie, Q.; Zhang, T.; and Ananiadou, S. 2024.
\newblock Emollms: A series of emotional large language models and annotation tools for comprehensive affective analysis.
\newblock In \emph{Proceedings of the 30th ACM SIGKDD Conference on Knowledge Discovery and Data Mining}, 5487--5496.

\bibitem[{Loshchilov and Hutter(2016)}]{loshchilov2016sgdr}
Loshchilov, I.; and Hutter, F. 2016.
\newblock Sgdr: Stochastic gradient descent with warm restarts.
\newblock \emph{arXiv preprint arXiv:1608.03983}.

\bibitem[{Meng et~al.(2022)Meng, Bau, Andonian, and Belinkov}]{meng2022locating}
Meng, K.; Bau, D.; Andonian, A.; and Belinkov, Y. 2022.
\newblock Locating and editing factual associations in gpt.
\newblock \emph{Advances in neural information processing systems}, 35: 17359--17372.

\bibitem[{Na et~al.(2025)Na, Hua, Wang, Shen, Yu, Wang, Wang, Torous, and Chen}]{na2025survey}
Na, H.; Hua, Y.; Wang, Z.; Shen, T.; Yu, B.; Wang, L.; Wang, W.; Torous, J.; and Chen, L. 2025.
\newblock A survey of large language models in psychotherapy: Current landscape and future directions.
\newblock \emph{arXiv preprint arXiv:2502.11095}.

\bibitem[{Park, Choe, and Veitch(2023)}]{park2023linear}
Park, K.; Choe, Y.~J.; and Veitch, V. 2023.
\newblock The linear representation hypothesis and the geometry of large language models.
\newblock \emph{arXiv preprint arXiv:2311.03658}.

\bibitem[{Poria et~al.(2018)Poria, Hazarika, Majumder, Naik, Cambria, and Mihalcea}]{poria2018meld}
Poria, S.; Hazarika, D.; Majumder, N.; Naik, G.; Cambria, E.; and Mihalcea, R. 2018.
\newblock Meld: A multimodal multi-party dataset for emotion recognition in conversations.
\newblock \emph{arXiv preprint arXiv:1810.02508}.

\bibitem[{Radford, Jozefowicz, and Sutskever(2018)}]{radford2018learning}
Radford, A.; Jozefowicz, R.; and Sutskever, I. 2018.
\newblock Learning to generate reviews and discovering sentiment.
\newblock \emph{arXiv preprint arXiv:1704.01444}.

\bibitem[{Rashkin et~al.(2018)Rashkin, Smith, Li, and Boureau}]{rashkin2018towards}
Rashkin, H.; Smith, E.~M.; Li, M.; and Boureau, Y.-L. 2018.
\newblock Towards empathetic open-domain conversation models: A new benchmark and dataset.
\newblock \emph{arXiv preprint arXiv:1811.00207}.

\bibitem[{Sabour et~al.(2024)Sabour, Liu, Zhang, Liu, Zhou, Sunaryo, Li, Lee, Mihalcea, and Huang}]{sabour2024emobench}
Sabour, S.; Liu, S.; Zhang, Z.; Liu, J.~M.; Zhou, J.; Sunaryo, A.~S.; Li, J.; Lee, T.; Mihalcea, R.; and Huang, M. 2024.
\newblock Emobench: Evaluating the emotional intelligence of large language models.
\newblock \emph{arXiv preprint arXiv:2402.12071}.

\bibitem[{Schlegel, Sommer, and Mortillaro(2025)}]{schlegel2025large}
Schlegel, K.; Sommer, N.~R.; and Mortillaro, M. 2025.
\newblock Large language models are proficient in solving and creating emotional intelligence tests.
\newblock \emph{Communications Psychology}, 3(1): 80.

\bibitem[{Shwartz-Ziv and Tishby(2017)}]{shwartz2017opening}
Shwartz-Ziv, R.; and Tishby, N. 2017.
\newblock Opening the black box of deep neural networks via information.
\newblock \emph{arXiv preprint arXiv:1703.00810}.

\bibitem[{Stade et~al.(2024)Stade, Stirman, Ungar, Boland, Schwartz, Yaden, Sedoc, DeRubeis, Willer, and Eichstaedt}]{stade2024large}
Stade, E.~C.; Stirman, S.~W.; Ungar, L.~H.; Boland, C.~L.; Schwartz, H.~A.; Yaden, D.~B.; Sedoc, J.; DeRubeis, R.~J.; Willer, R.; and Eichstaedt, J.~C. 2024.
\newblock Large language models could change the future of behavioral healthcare: a proposal for responsible development and evaluation.
\newblock \emph{NPJ Mental Health Research}, 3(1): 12.

\bibitem[{Tak et~al.(2025)Tak, Banayeeanzade, Bolourani, Kian, Jia, and Gratch}]{tak2025mechanistic}
Tak, A.~N.; Banayeeanzade, A.; Bolourani, A.; Kian, M.; Jia, R.; and Gratch, J. 2025.
\newblock Mechanistic Interpretability of Emotion Inference in Large Language Models.
\newblock \emph{arXiv preprint arXiv:2502.05489}.

\bibitem[{Tak and Gratch(2024)}]{tak2024gpt}
Tak, A.~N.; and Gratch, J. 2024.
\newblock Gpt-4 emulates average-human emotional cognition from a third-person perspective.
\newblock In \emph{2024 12th International Conference on Affective Computing and Intelligent Interaction (ACII)}, 337--345. IEEE.

\bibitem[{Tenney, Das, and Pavlick(2019)}]{tenney2019bert}
Tenney, I.; Das, D.; and Pavlick, E. 2019.
\newblock BERT rediscovers the classical NLP pipeline.
\newblock \emph{arXiv preprint arXiv:1905.05950}.

\bibitem[{Vzorinab et~al.(2024)Vzorinab, Bukinichac, Sedykha, Vetrovab, and Sergienkob}]{vzorinab2024emotional}
Vzorinab, G.~D.; Bukinichac, A.~M.; Sedykha, A.~V.; Vetrovab, I.~I.; and Sergienkob, E.~A. 2024.
\newblock The emotional intelligence of the GPT-4 large language model.
\newblock \emph{Psychology in Russia: State of the art}, 17(2): 85--99.

\bibitem[{Wang et~al.(2024)Wang, Li, Chang, Wang, and Wu}]{wang2024negativeprompt}
Wang, X.; Li, C.; Chang, Y.; Wang, J.; and Wu, Y. 2024.
\newblock Negativeprompt: Leveraging psychology for large language models enhancement via negative emotional stimuli.
\newblock \emph{arXiv preprint arXiv:2405.02814}.

\bibitem[{Wei and Zou(2019)}]{wei2019eda}
Wei, J.; and Zou, K. 2019.
\newblock EDA: Easy data augmentation techniques for boosting performance on text classification tasks.
\newblock \emph{arXiv preprint arXiv:1901.11196}.

\bibitem[{wenknow(2025)}]{wenknow2025datauniversereddit_dataset_888}
wenknow. 2025.
\newblock The Data Universe Datasets: The finest collection of social media data the web has to offer.

\bibitem[{Yang et~al.(2025)Yang, Li, Yang, Zhang, Hui, Zheng, Yu, Gao, Huang, Lv et~al.}]{yang2025qwen3}
Yang, A.; Li, A.; Yang, B.; Zhang, B.; Hui, B.; Zheng, B.; Yu, B.; Gao, C.; Huang, C.; Lv, C.; et~al. 2025.
\newblock Qwen3 technical report.
\newblock \emph{arXiv preprint arXiv:2505.09388}.

\bibitem[{Zhang et~al.(2025{\natexlab{a}})Zhang, Zhang, Sun, Zhao, and Jin}]{zhang2025exploring}
Zhang, X.; Zhang, T.; Sun, L.; Zhao, J.; and Jin, Q. 2025{\natexlab{a}}.
\newblock Exploring interpretability in deep learning for affective computing: a comprehensive review.
\newblock \emph{ACM Transactions on Multimedia Computing, Communications and Applications}.

\bibitem[{Zhang et~al.(2025{\natexlab{b}})Zhang, Wang, Wu, Tiwari, Li, Wang, and Qin}]{zhang2025dialoguellm}
Zhang, Y.; Wang, M.; Wu, Y.; Tiwari, P.; Li, Q.; Wang, B.; and Qin, J. 2025{\natexlab{b}}.
\newblock Dialoguellm: Context and emotion knowledge-tuned large language models for emotion recognition in conversations.
\newblock \emph{Neural Networks}, 107901.

\bibitem[{Zhang et~al.(2024)Zhang, Yang, Xu, Gao, Huang, Mu, Feng, Wang, Zhang, Song et~al.}]{zhang2024affective}
Zhang, Y.; Yang, X.; Xu, X.; Gao, Z.; Huang, Y.; Mu, S.; Feng, S.; Wang, D.; Zhang, Y.; Song, K.; et~al. 2024.
\newblock Affective computing in the era of large language models: A survey from the nlp perspective.
\newblock \emph{arXiv preprint arXiv:2408.04638}.

\bibitem[{Zhao et~al.(2024)Zhao, Okawa, Bigelow, Yu, Ullman, and Tanaka}]{zhao2024emergence}
Zhao, B.; Okawa, M.; Bigelow, E.~J.; Yu, R.; Ullman, T.; and Tanaka, H. 2024.
\newblock Emergence of hierarchical emotion representations in large language models.

\end{thebibliography}

\end{document}